\documentclass[conference]{IEEEtran}
\IEEEoverridecommandlockouts

\usepackage{graphicx}
\usepackage{subcaption}
\usepackage{svg}

\usepackage{amsmath}
\usepackage{amsfonts}

\usepackage{algorithmic}

\usepackage{array}

\ifCLASSOPTIONcompsoc
 \usepackage[caption=false,font=normalsize,labelfont=sf,textfont=sf]{subfig}
\else
 \usepackage[caption=false,font=footnotesize]{subfig}
\fi
\usepackage{stfloats}

\usepackage{url}
\usepackage{xcolor}
\def\BibTeX{{\rm B\kern-.05em{\sc i\kern-.025em b}\kern-.08em
    T\kern-.1667em\lower.7ex\hbox{E}\kern-.125emX}}
\begin{document}

\title{Large Language Models (LLMs) Assisted Wireless Network Deployment in Urban Settings}

\author{\IEEEauthorblockN{Nurullah Sevim}
\IEEEauthorblockA{\textit{Electrical and Computer Engineering}\\
\textit{Texas A\&M University}\\
College Station, Texas, USA\\
Email: nurullahsevim@tamu.edu}
\and
\IEEEauthorblockN{Mostafa Ibrahim}
\IEEEauthorblockA{\textit{Electrical and Computer Engineering}\\
\textit{Texas A\&M University}\\
College Station, Texas, USA\\
Email: mostafa.ibrahim@tamu.edu}
\and
\IEEEauthorblockN{Sabit Ekin}
\IEEEauthorblockA{
\textit{Engineering Technology, }\\
\textit{and Electrical and Computer Engineering}\\
\textit{Texas A\&M University}\\
College Station, Texas, USA\\
Email: sabitekin@tamu.edu}}

\maketitle

\begin{abstract}

The advent of Large Language Models (LLMs) has revolutionized language understanding and human-like text generation, drawing interest from many other fields with this question in mind: What else are the LLMs capable of? Despite their widespread adoption, ongoing research continues to explore new ways to integrate LLMs into diverse systems.

This paper explores new techniques to harness the power of LLMs for 6G (6th Generation) wireless communication technologies, a domain where automation and intelligent systems are pivotal. The inherent adaptability of LLMs to domain-specific tasks positions them as prime candidates for enhancing wireless systems in the 6G landscape.

We introduce a novel Reinforcement Learning (RL) based framework that leverages LLMs for network deployment in wireless communications. Our approach involves training an RL agent, utilizing LLMs as its core, in an urban setting to maximize coverage. The agent's objective is to navigate the complexities of urban environments and identify the network parameters for optimal area coverage. Additionally, we integrate LLMs with Convolutional Neural Networks (CNNs) to capitalize on their strengths while mitigating their limitations. The Deep Deterministic Policy Gradient (DDPG) algorithm is employed for training purposes. The results suggest that LLM-assisted models can outperform CNN-based models in some cases while performing at least as well in others.

\end{abstract}

%
\IEEEpeerreviewmaketitle

\section{Introduction}

Large Language Models (LLMs) like GPT \cite{brown2020language}, BERT \cite{Devlin2018}, and their successors have significantly impacted Natural Language Processing (NLP) by excelling across various language understanding and generation benchmarks. The transfer learning capabilities of LLMs facilitate easy adaptation to various fields at minimal cost. This adaptability involves fine-tuning pre-trained models like DistillBERT \cite{sanh2019distilbert} on specific datasets, allowing the models to efficiently apply their broad knowledge to new tasks. This process has not only demonstrated the effectiveness of LLMs but also their flexibility across different domains, encouraging researchers to explore and utilize LLM-based solutions extensively.

Wireless communications is another field that can utilize LLMs extensively in its applications. In the era of 6G, time variations in wireless links makes it essential to have decision-makers that are quick, strong, and dependable. Using LLMs to automate network systems presents a practical approach to develop these types of decision-makers.

Wireless network deployment contains many challenges, including the design and maintenance of robust connections, efficient routing in dynamic environments, and energy optimization \cite{Goldsmith2005}. Especially, some ad hoc wireless networks without fixed infrastructure need smart design solutions to ensure a consistent and high-quality wireless links. The dynamic nature of these networks can also require adaptive planning that can manage mobility and variable link quality \cite{Rappaport2002}. Besides, energy efficiency is critical, especially in battery-powered networks, where there is a trade-off between the life span of the network and network performance.

In network deployment scenarios, two main concepts to consider are the path loss and shadowing effect. The wireless systems are typically engineered with a target minimum received power \(P_{\min}\), which is crucial for acceptable performance. For instance, failing to meet this threshold in cellular networks can cause significant degradation in received signal thus lower data rates. Due to shadowing, the received power at any distance from the transmitter is log-normally distributed, posing a probability of falling below \(P_{\min}\). Thus, the effects of path loss and shadowing necessitate careful planning and deployment strategies for wireless networks to establish consistent and high-quality communication links, hence minimizing coverage holes.

Self-organizing networks, a well-studied topic in the literature \cite{sonsurvey}, introduces a way for managing and optimizing network performance autonomously. Exploiting previous experiences, such networks can autonomously arrange their location and network configurations to increase service quality  \cite{mullany05son}. These networks utilize algorithms and protocols to dynamically adjust to changes in the environment aiming to increase efficiency, reliability, and scalability \cite{aliu13son}. Self-organizing networks are presented as a suitable candidate for building ad hoc networks with dynamic structure and low-powered nodes \cite{ho03son}. 

In this paper,  we introduce a novel approach that integrates LLMs into the process of wireless network deployment, where we use Reinforcement Learning (RL) to train an LLM-assisted agent. The objective is to determine the optimal location  and orientation for a base station using the LLM-assisted RL agent in an urban setting to maximize the coverage. To achieve this, we employ an advanced actor-critic RL methodology known as Deep Deterministic Policy Gradients (DDPG) \cite{lillicrap2015continuous} for the training of our agent. Our framework incorporates LLMs in two distinct configurations: first, we integrate an LLM, specifically DistilBERT, within the actor network model; second, we combine LLM with Convolutional Neural Networks (CNNs) to increase its effectiveness. We use a third configuration where the actor network consists only of CNNs, without LLM, for comparison purposes.

We justify using LLMs instead of other general-purpose neural networks by the following reasoning: general neural networks primarily utilize loss functions to understand their objectives. A significant advantage of leveraging LLMs lies in their ability to process textual inputs, also known as prompts. This capability allows LLMs to receive more explicit and detailed descriptions of their objectives and a more comprehensive explanation of the environment in which they operate, as opposed to solely depending on loss functions for guidance. Moreover, LLMs can be efficiently directed through immediate prompts, making them well-suited for rapidly changing environments.

Besides, there are many publicly available pre-trained LLMs which are trained on extensive textual datasets encompassing a wide range of domains \cite{sanh2019distilbert,roberta}, including wireless communications. This diversity in training material enables these models to possess a broad knowledge base. Our objective is to reveal the latent information embedded within LLMs and leverage this information to develop decision-making mechanisms based on LLMs.

We use Sionna library \cite{sionna}\footnote{Sionna™ is an open-source Python library for link-level simulations of digital communication systems built on top of the open-source software library TensorFlow for machine learning. The official documentation can be found in this link: https://nvlabs.github.io/sionna/.} in Python to set the wireless environment for our experiments. We devised multiple scenarios to comparatively analyze the performance of our agent with different configurations. 

The rest of the paper is organized as follows: in Section \ref{pre}, some preliminary background is provided for several relevant topics. In Section \ref{method} the environment settings and network architectures are extensively explained. Experimental details, results, and discussions are presented in Section \ref{exp}. Finally, conclusions are drawn in Section \ref{conc}.

\section{Preliminaries}
\label{pre}

\subsection{Path Loss}

Path loss is the attenuation a signal experiences as it propagates through space. It is primarily due to the spreading of the wavefront, which causes the signal power to decrease with distance. The free-space path loss (FSPL) model is given by in terms of decibels (dB):

\begin{equation}
    \text{FSPL (dB)} = 20\log_{10}(d) + 20\log_{10}(f) + 20\log_{10}\left(\frac{4\pi}{c}\right),
\end{equation}
where \(d\) is the distance between the transmitter and receiver in meters,
     \(f\) is the frequency of the signal in Hertz,
     \(c\) is the speed of light (\(3 \times 10^8\) m/s).

\subsection{Shadowing}

Shadowing, or slow fading, occurs when obstacles in the environment, such as buildings or trees, block the direct signal path between the transmitter and receiver. It is modeled using a log-normal distribution, where the signal power in dBm is normally distributed around the mean path loss with a standard deviation dependent on the environment. The model can be expressed as:

\begin{equation}
    P_r(dBm) = P_t(dBm) - PL(dB) - X_\sigma(dB),
\end{equation}
where \(P_r(dBm)\) is the received power in dBm,
     \(P_t(dBm)\) is the transmitted power in dBm,
     \(PL(dB)\) is the path loss in dB calculated using an appropriate path loss model,
     \(X_\sigma(dB)\) is a zero-mean Gaussian random variable with standard deviation \(\sigma\), representing shadowing effects.

\subsection{Deep Deterministic Policy Gradients (DDPG)}
\label{ddpg}

The DDPG algorithm is an actor-critic method tailored for continuous action spaces using deep learning. Introduced by \cite{lillicrap2015continuous}, DDPG blends policy gradient and Q-learning methods to manage high-dimensional state information from the environment. It consists of two primary components: an actor, which determines actions based on the current state, and a critic, which evaluates these actions by predicting their potential rewards.

DDPG stabilizes training through experience replay and target networks for both actor and critic. Experience replay mitigates temporal correlation by storing transitions (state, action, reward, next state) in a buffer and sampling randomly for training, which enhances learning efficiency. Target networks provide stable, consistent targets during value updates, contributing to the overall stability of the training process.

DDPG is notable for its efficacy in continuous action spaces, applicable in fields like robotics and complex video games where actions cannot be discretized effectively. This attribute is crucial for our experiments requiring continuous responses from the agent.

The critic network computes the action-value function \(Q(s, a)\), indicating the expected return for an action \(a\) in state \(s\), following the current policy. The critic updates by minimizing the loss:

\begin{equation}
L(\theta^Q) = \mathbb{E}_{s,a,r,s'}\left[\left(Q(s, a | \theta^Q) - y\right)^2\right],
\end{equation}
where \(y = r + \gamma \Bar{Q}(s', \mu'(s' | \theta^{\mu'}) | \theta^{\Bar{Q}})\), with \(\Bar{Q}\) as the critic's target network action-value function, \(\theta\) as the network weights, \(\mu'\) as the actor's target network decision, \(r\) as the reward, \(\gamma\) as the discount factor, and \(s'\) as the subsequent state.

The actor network updates its policy using the gradient of the expected return from the initial state distribution \(J\) concerning the actor parameters \(\theta^\mu\):
\begin{equation}
\nabla_{\theta^\mu} J \approx \mathbb{E}_{s}\left[\nabla_a Q(s, a | \theta^Q)|_{a=\mu(s)} \nabla_{\theta^\mu} \mu(s | \theta^\mu)\right].
\end{equation}

DDPG incorporates an exploration strategy using the Ornstein-Uhlenbeck process \cite{ornstein}, defined as:
\begin{equation}
    dX_t = \theta (\mu - X_t) dt + \sigma dW_t,
\end{equation}
where \(X_t\) is the process value at time \(t\), \(\theta\) is a rate parameter, \(\mu\) is the mean, \(\sigma\) is the volatility parameter, and \(dW_t\) is the increment of a Wiener process. This noise is added to the deterministic policy actions to promote exploration:
\begin{equation}
    a'_t = \mu(s_t | \theta^\mu) + \mathcal{N}_t,
\end{equation}
where \(\mu(s_t | \theta^\mu)\) is the deterministic policy function, and \(\mathcal{N}_t\) is the noise sampled from the Ornstein-Uhlenbeck process.

\begin{figure}
   \centering
   \includegraphics[width=0.75\linewidth]{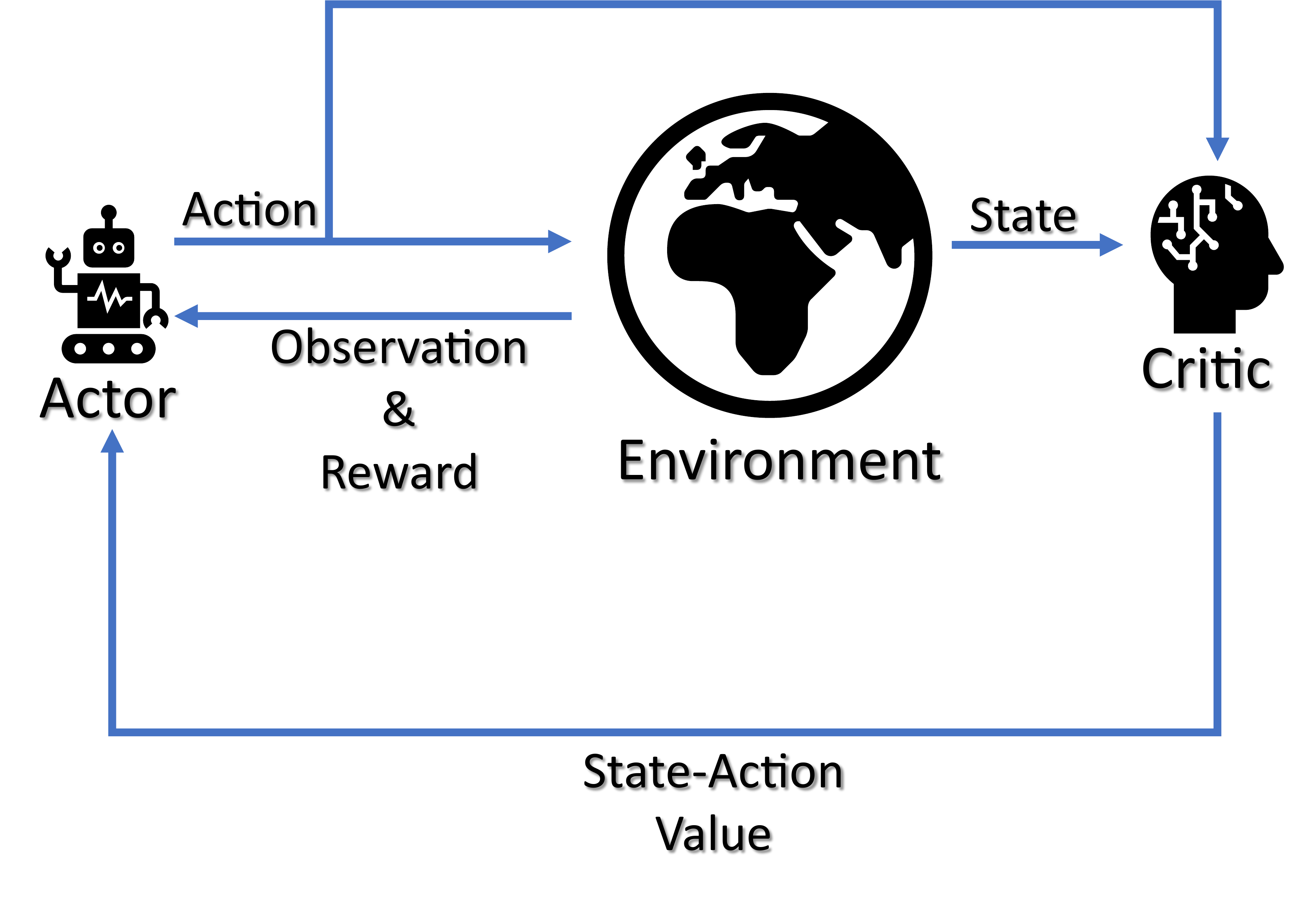}
   \caption{A conceptual diagram of the DDPG algorithm, showcasing the interactions between the actor and critic networks, the policy behavior, and the value functions.}
   \label{RL}
\end{figure}

\section{Methodology}
\label{method}

To deploy LLMs in urban wireless communication settings, we explore various scenarios to strategically position base stations, aiming to maximize signal strength for users at specific locations. Rather than engaging in complex mathematical analysis of the environment's electromagnetic properties, we leverage LLMs to interpret these characteristics through a learning process. Sionna's ray tracing technology, which simulates how light interacts with objects to calculate effects like reflections and shadows, supports this.

We use this approach to identify optimal locations and orientations for base stations in urban areas using multiple scenarios. Additionally, we have implemented a fine-tuning process for the pre-trained DistilBERT model using the DDPG actor-critic Reinforcement Learning algorithm. In this setup, the LLM functions as a decision-making agent, optimizing signal strength at targeted locations based on specifically designed prompts.

\subsection{System Settings}

\begin{figure}
    \begin{subfigure}[t]{0.5\textwidth}
        \centering
        \includegraphics[width=0.75\textwidth]{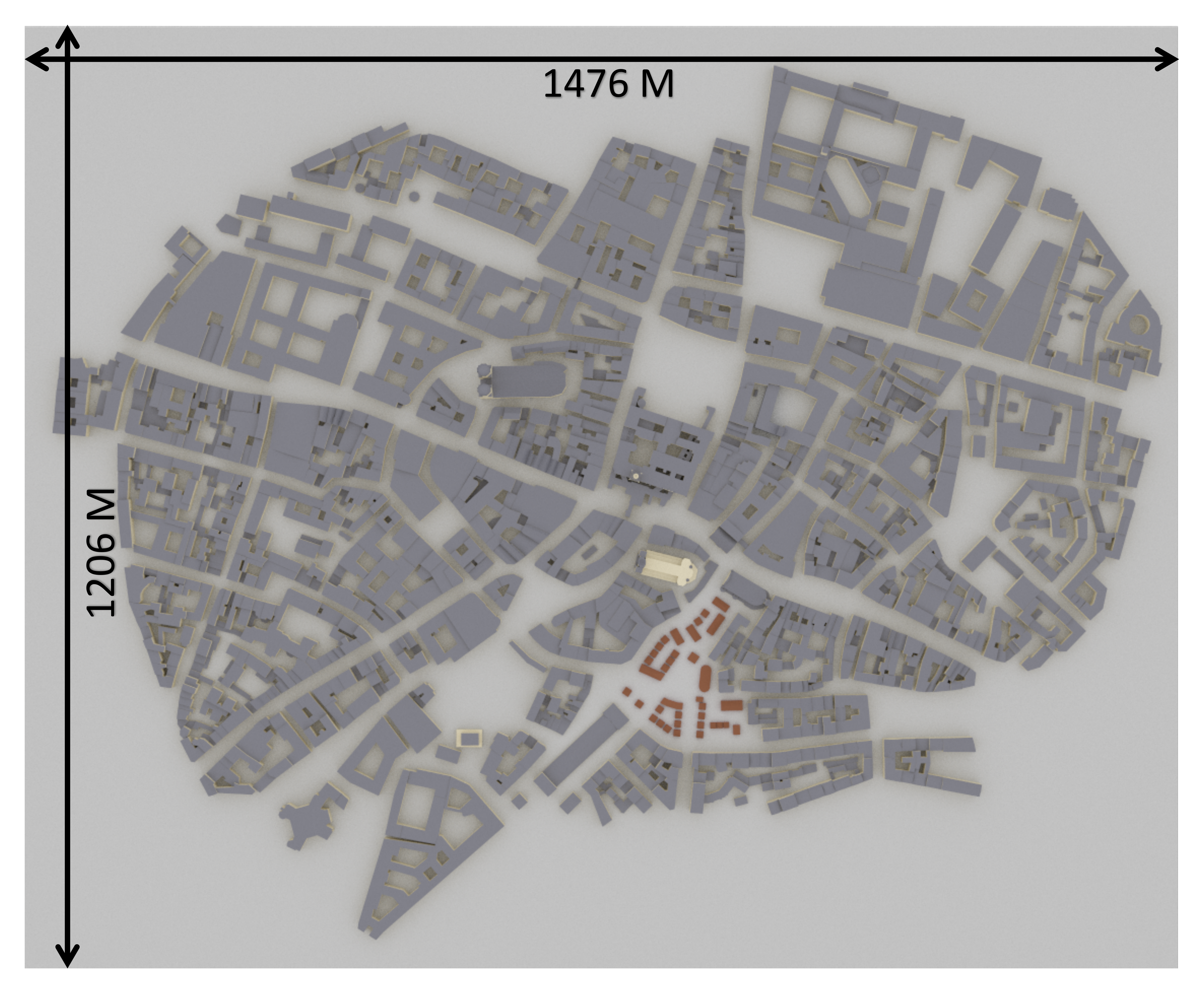}
        \caption{Bird's-eye view.}
        \label{scenes-bird}
    \end{subfigure}
    ~ 
    \begin{subfigure}[t]{0.5\textwidth}
        \centering
        \includegraphics[width=0.75\textwidth]{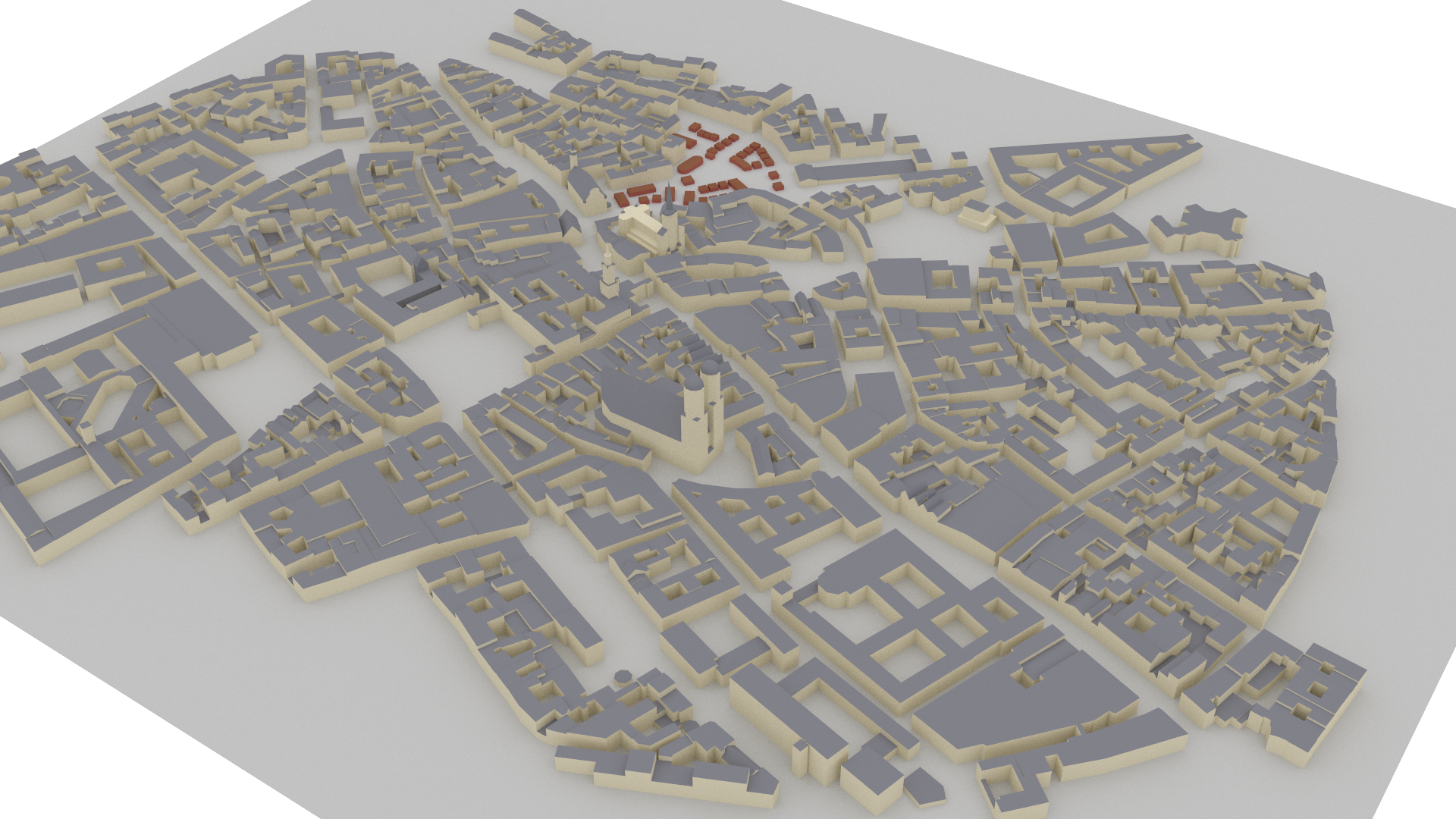}
        \caption{3D view.}
    \end{subfigure}    
    \caption{The 'Munich' environment from different angles.}
\label{scenes}
\end{figure}

Before explaining the details of the algorithms and how we use the LLMs as decision maker, we first need to understand the utilized environment and the considered scenarios. We utilize the Ray Tracing module of Sionna \cite{sionna} library to setup the environment for our experiments. To fully understand how this module works and how it simulates wireless environments, we refer the reader to the Sionna's documentations\footnote{https://nvlabs.github.io/sionna/api/rt.html}. We choose to work on one of the default 3D environments exists in Sionna library, called 'Munich'. In Fig. \ref{scenes}, the pictures of the `Munich' environment are given from bird's-eye view and 3D view.

As shown in Fig. \ref{scenes}, the depicted environment is an urban setting, characterized by the presence of numerous buildings. In the context of wireless communication, this dense urban architecture implies that the transmitted signal is prone to obstructions. Consequently, this complexity renders the identification of an explicit solution for the stated objective unfeasible.

\begin{figure*}
    \centering
    \begin{subfigure}[t]{0.32\textwidth}
        \centering
        \includegraphics[width=0.9\textwidth]{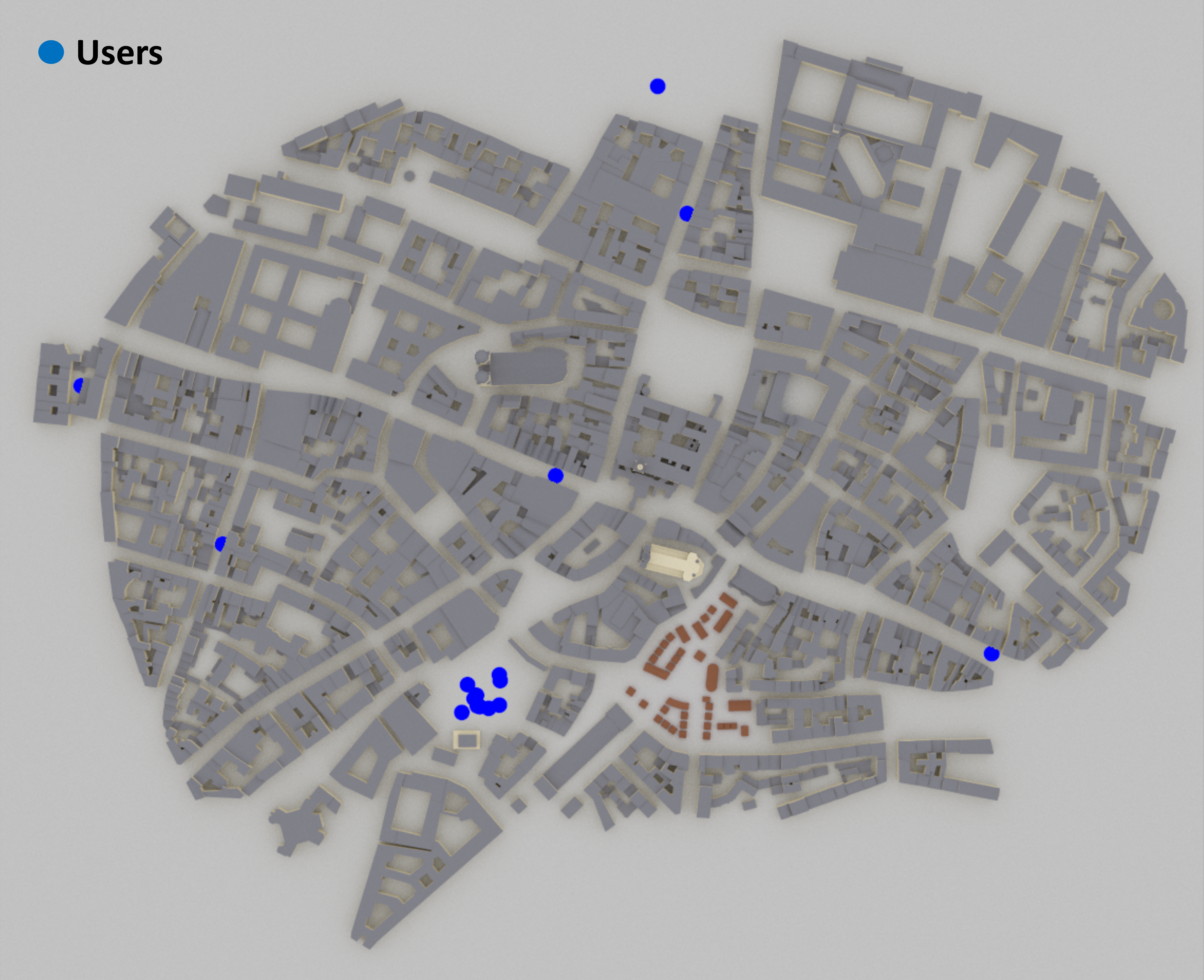}
        \caption{Case-1}
    \end{subfigure}%
    ~ 
    \begin{subfigure}[t]{0.32\textwidth}
        \centering
        \includegraphics[width=0.9\textwidth]{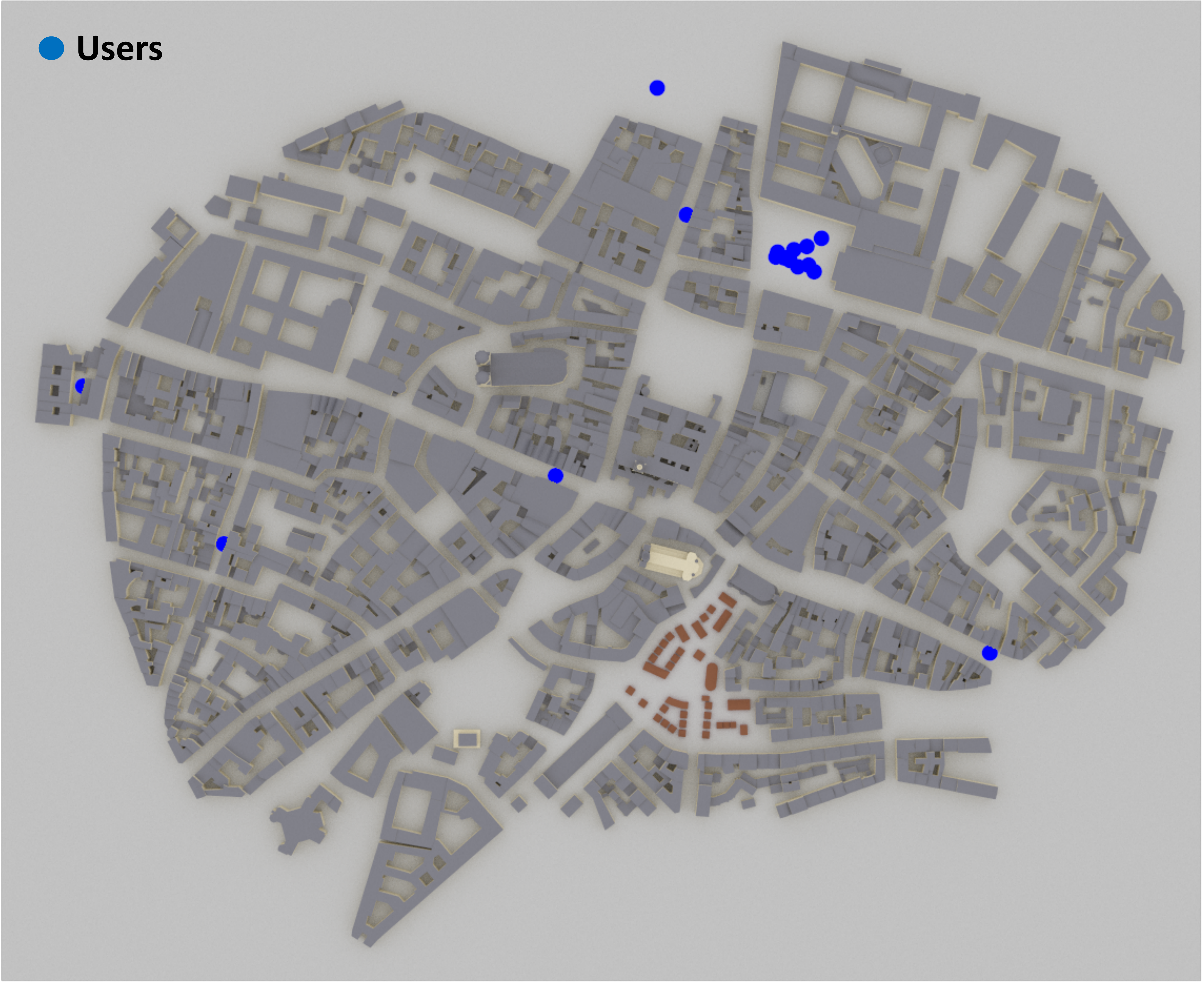}
        \caption{Case-2}
    \end{subfigure}%
    ~ 
    \begin{subfigure}[t]{0.32\textwidth}
        \centering
        \includegraphics[width=0.9\textwidth]{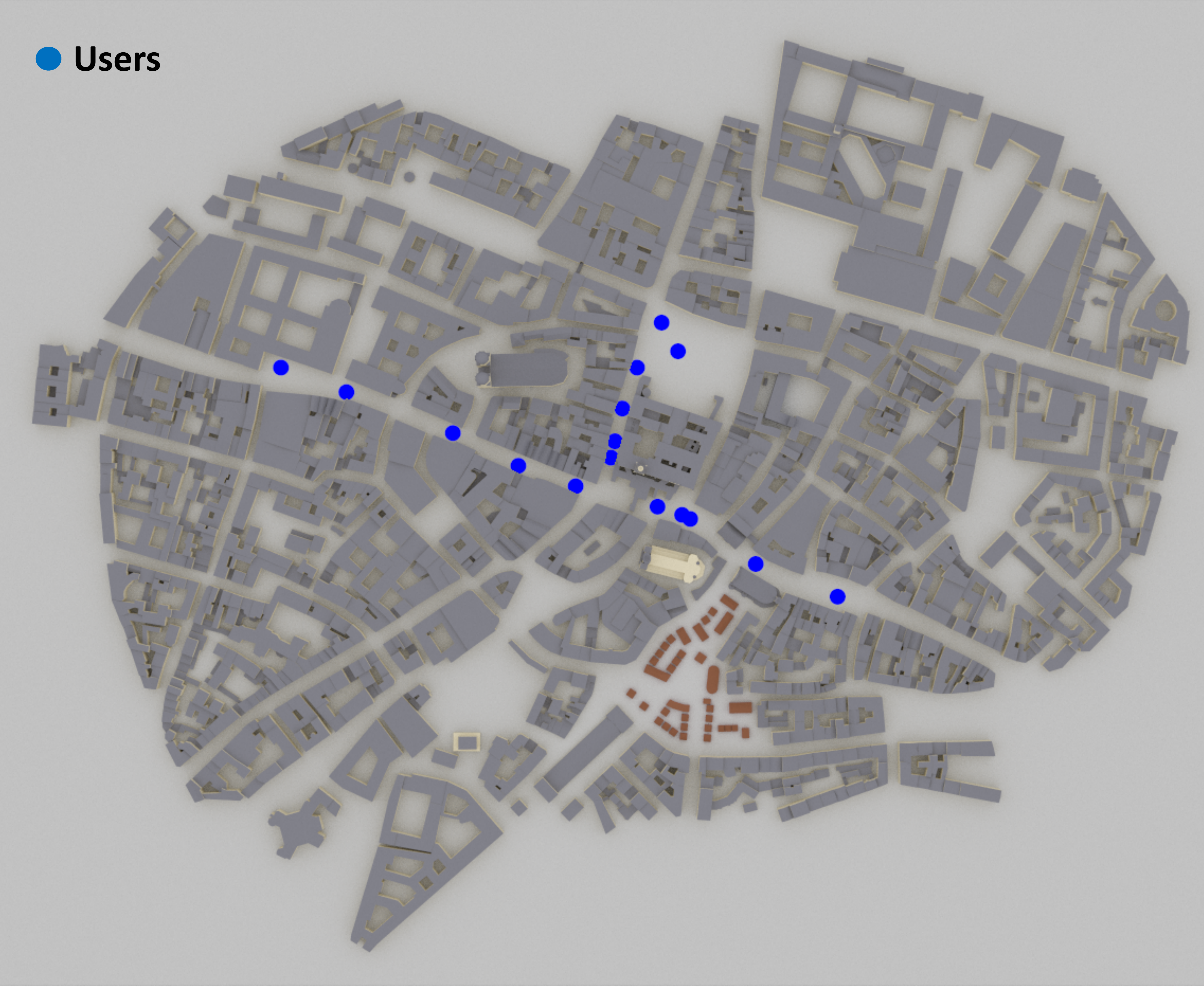}
        \caption{Case-3}
    \end{subfigure}%
    
    \caption{In all of these scenarios, we want all the users to get the maximal signal reception.}
\label{cases}
\end{figure*}

For the experiment, three distinct scenarios were devised to simulate conditions where a crowd is present in a specific area on the map or there is a traffic jam. The goal is to adjust the location and orientation of a base station in such a way as to optimally serve the users. Fig. \ref{cases} illustrates the three scenarios that we set up. In the first two cases, we simulate crowds in two different open areas on the map. In the third case, a traffic jam on the main roads is simulated.

\begin{figure}
    \centering
    \includegraphics[width=0.9\linewidth]{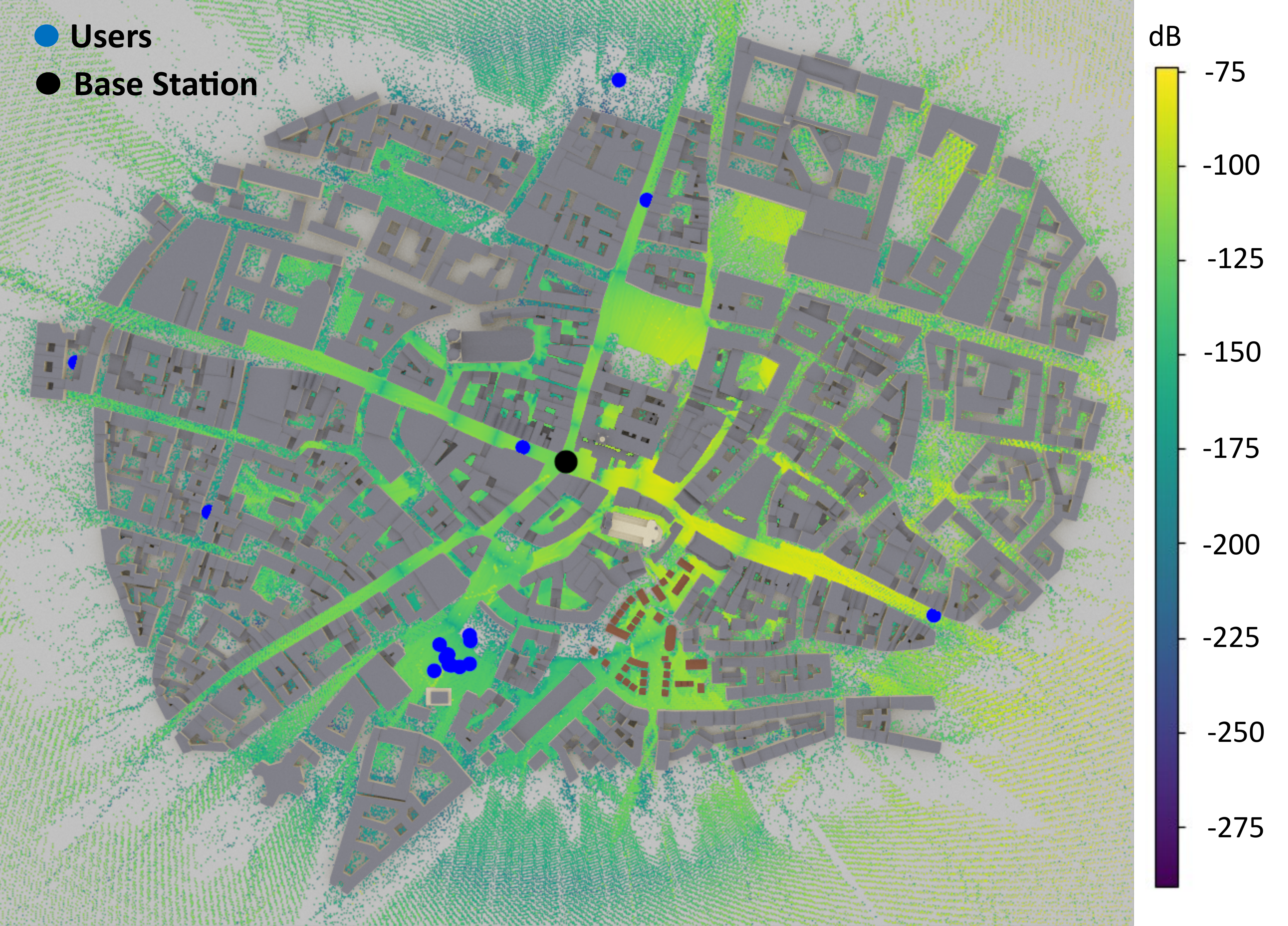}
    \caption{A visualization of the coverage map that is obtained during the training.}
    \label{coverage-map}
\end{figure}

In all of these scenarios, we manage the base station to meet the objective by configuring the 3D location and orientation of it. The Sionna library provides the ability to calculate the coverage for the entire map. The received signal powers for the specified users are acquired from the coverage map. One example of the coverage map is given in Fig. \ref{coverage-map}. The carrier frequency used in all these scenarios is 3.9 GHz.

Next, to calculate the received signal strength at the user end, we assume the base station has explicit planar antenna array and the users all have a dipole antenna. The configuration for the antenna arrays are given in Table \ref{array-config}. In the table, `tr38901' stands for array pattern described in Technical Report (TR) number 38.901 from 3GPP documentation.

\begin{table}[]
\caption{The configuration for antenna arrays of base station (Transmitter) and user equipment (Receiver).}
\label{array-config}
\resizebox{\linewidth}{!}{%
\begin{tabular}{c|c|c|c|c|c|c|}
\cline{2-7}
                                                                                                       & \textbf{\begin{tabular}[c]{@{}c@{}}\# of \\ rows\end{tabular}} & \textbf{\begin{tabular}[c]{@{}c@{}}\# of \\ columns\end{tabular}} & \textbf{\begin{tabular}[c]{@{}c@{}}Vertical \\ Spacing\end{tabular}} & \textbf{\begin{tabular}[c]{@{}c@{}}Horizontal \\ Spacing\end{tabular}} & \textbf{Pattern} & \textbf{Polarization} \\ \hline
\multicolumn{1}{|c|}{\textbf{\begin{tabular}[c]{@{}c@{}}Transmitter \\ Antenna \\ Array\end{tabular}}} & 8                                                              & 2                                                                 & 0.7                                                                  & 0.5                                                                    & tr38901          & VH                    \\ \hline
\multicolumn{1}{|c|}{\textbf{\begin{tabular}[c]{@{}c@{}}Receiver \\ Antenna \end{tabular}}}    & 1                                                              & 1                                                                 & -                                                                    & -                                                                      & Dipole           & Cross                 \\ \hline
\end{tabular}%
}
\end{table}

\subsection{Actor-Critic Network Settings}

Fine-tuning is a crucial process to adapt LLMs to work on different tasks effectively. Here, we use DDPG algorithm to fine-tune the LLM. In DDPG, the function approximators in actor and critic are comprise of deep neural networks models, where we used several different architectures. 

\begin{figure}
    \centering
    \includegraphics[width=0.8\linewidth]{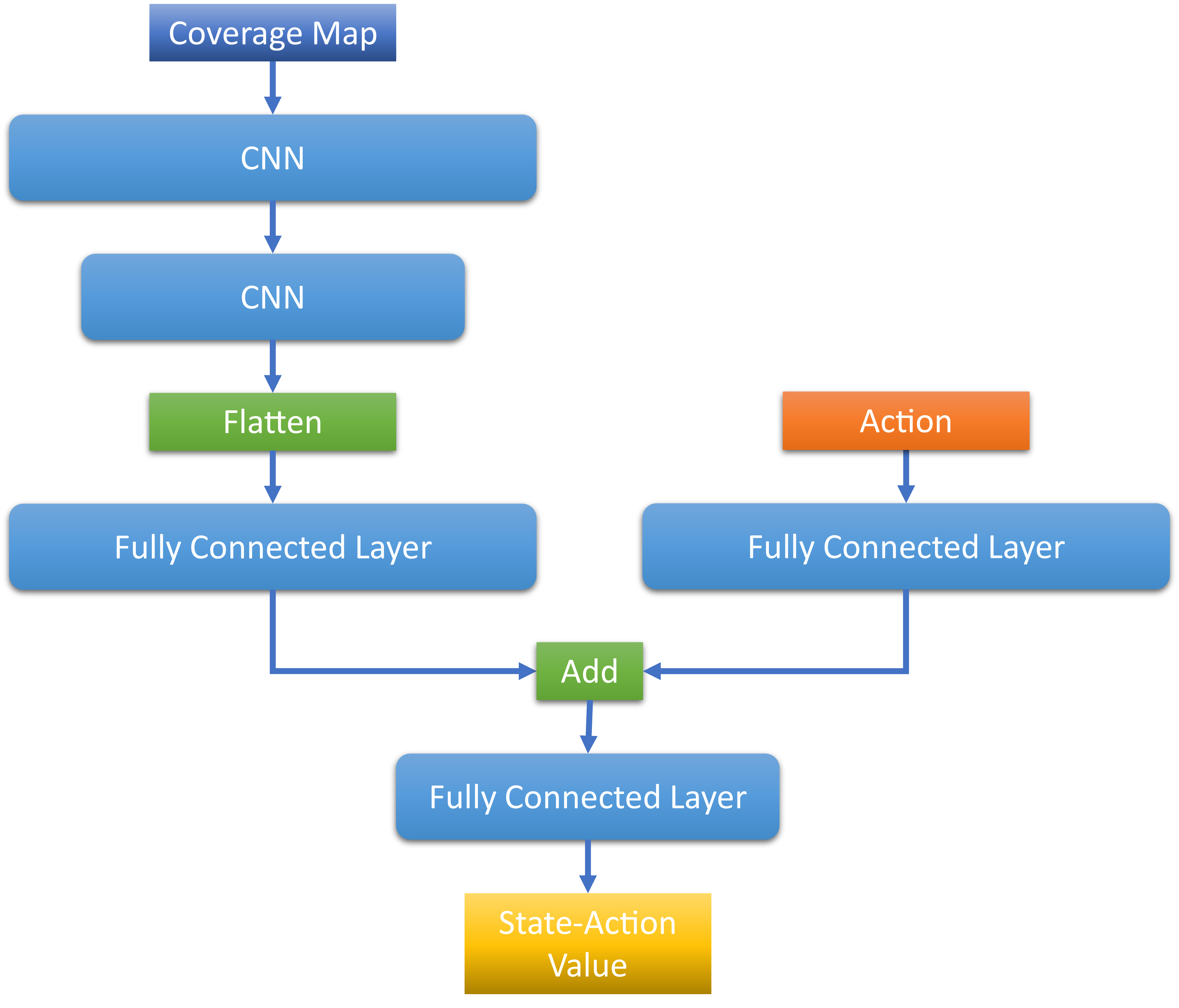}
    \caption{The network architecture of critic model in DDPG is depicted. The left branch processes the coverage map as the state observation with consecutive CNN layers. The reason of using two consecutive CNN layers with reducing sizes is to avoid having too many parameters in the following fully connected layer. The right branch processes the action decided by the actor model. Both branches use fully connected layers to return outputs in the same size, which in our case is 256. The outputs of both branches are added up and passed through another fully connected layer to give a scalar State-Action Value.}
    \label{critic}
\end{figure}

The critic network uses a coverage map, a 1206 by 1476 matrix, where each element measures received signal strength in dB for each unit square. This map is processed using CNNs, detailed in Fig. \ref{critic}. The critic outputs a scalar indicating the quality of actions taken by the actor, based on the current state represented by the coverage map. The actions pertain to the 3D location $(x,y,z)$ and orientation $(\alpha,\theta,\phi)$ of the base station, where $x,y\in[-500,500]$, $z\in[20,120]$, $\alpha\in[-\pi,\pi]$, $\theta\in[-\pi/2,\pi/2]$, and $\phi\in[-\pi,\pi]$.

\begin{figure*}
    \centering
    \begin{subfigure}{0.33\textwidth}
        \centering
        \includegraphics[width=0.6\textwidth]{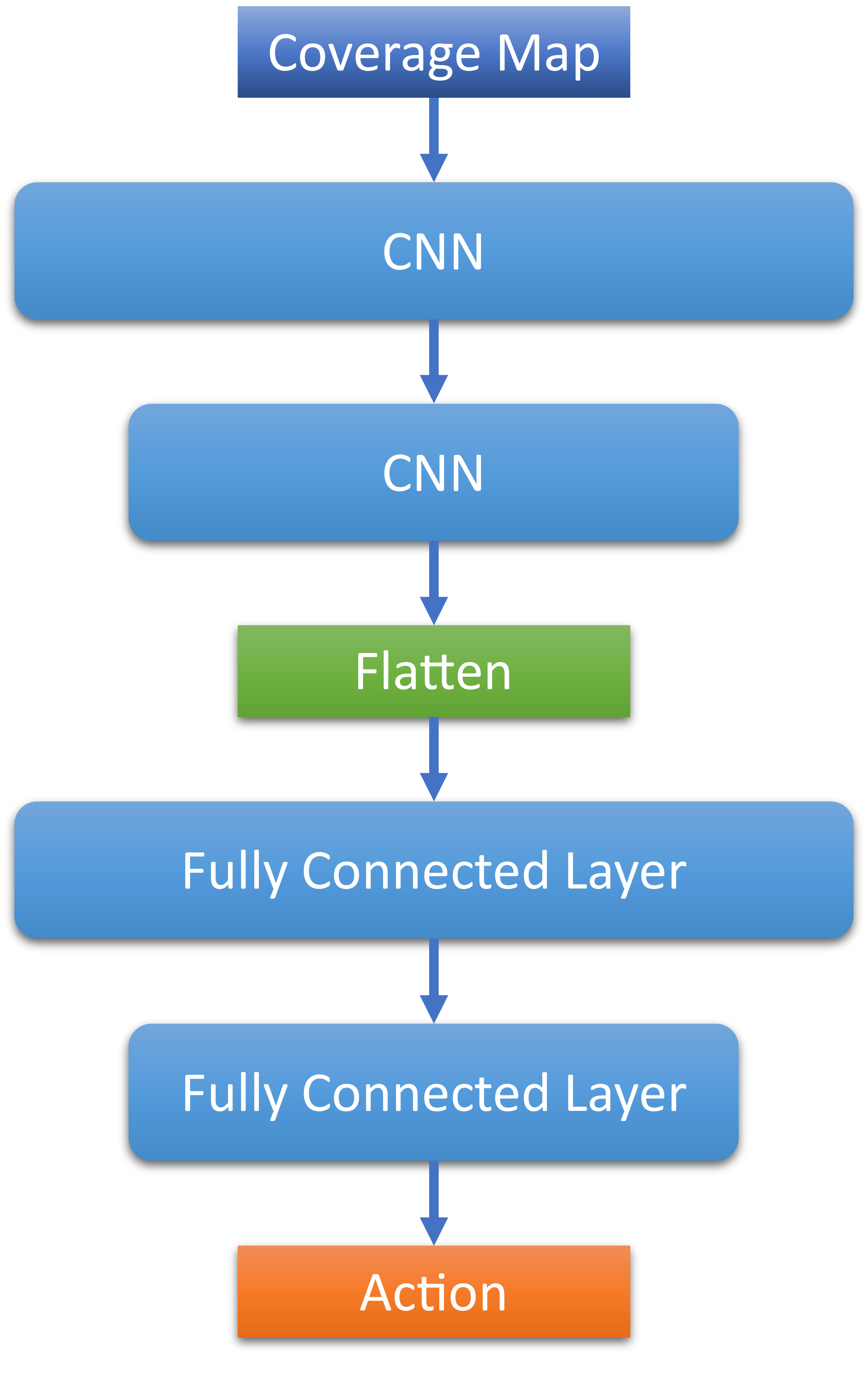}
        \caption{CNN-Only Actor}
    \end{subfigure}%
    ~
    \begin{subfigure}{0.33\textwidth}
        \centering
        \includegraphics[width=0.6\textwidth]{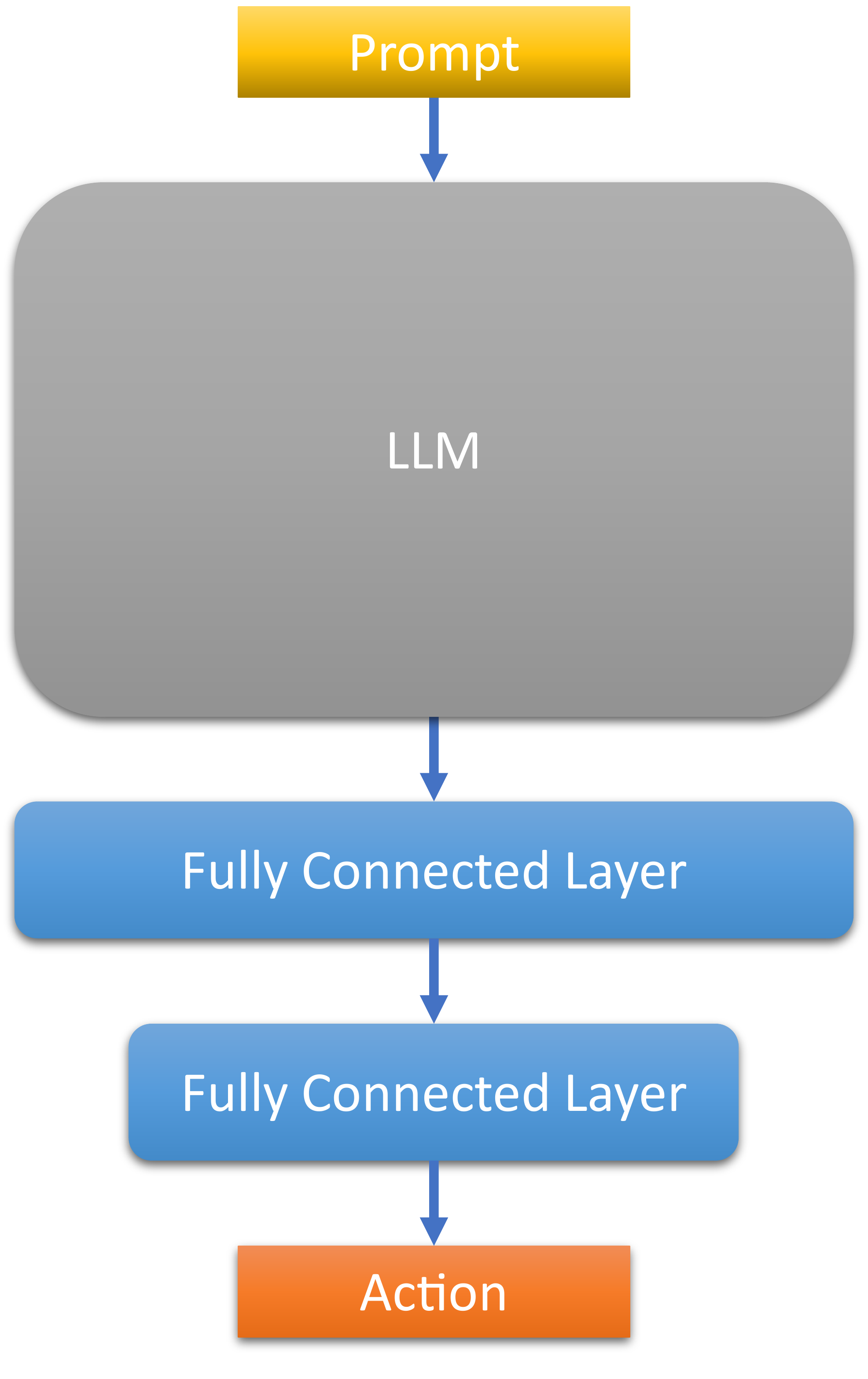}
        \caption{LLM-Only Actor}
    \end{subfigure}%
    ~
    \begin{subfigure}{0.33\textwidth}
        \centering
        \includegraphics[width=\textwidth]{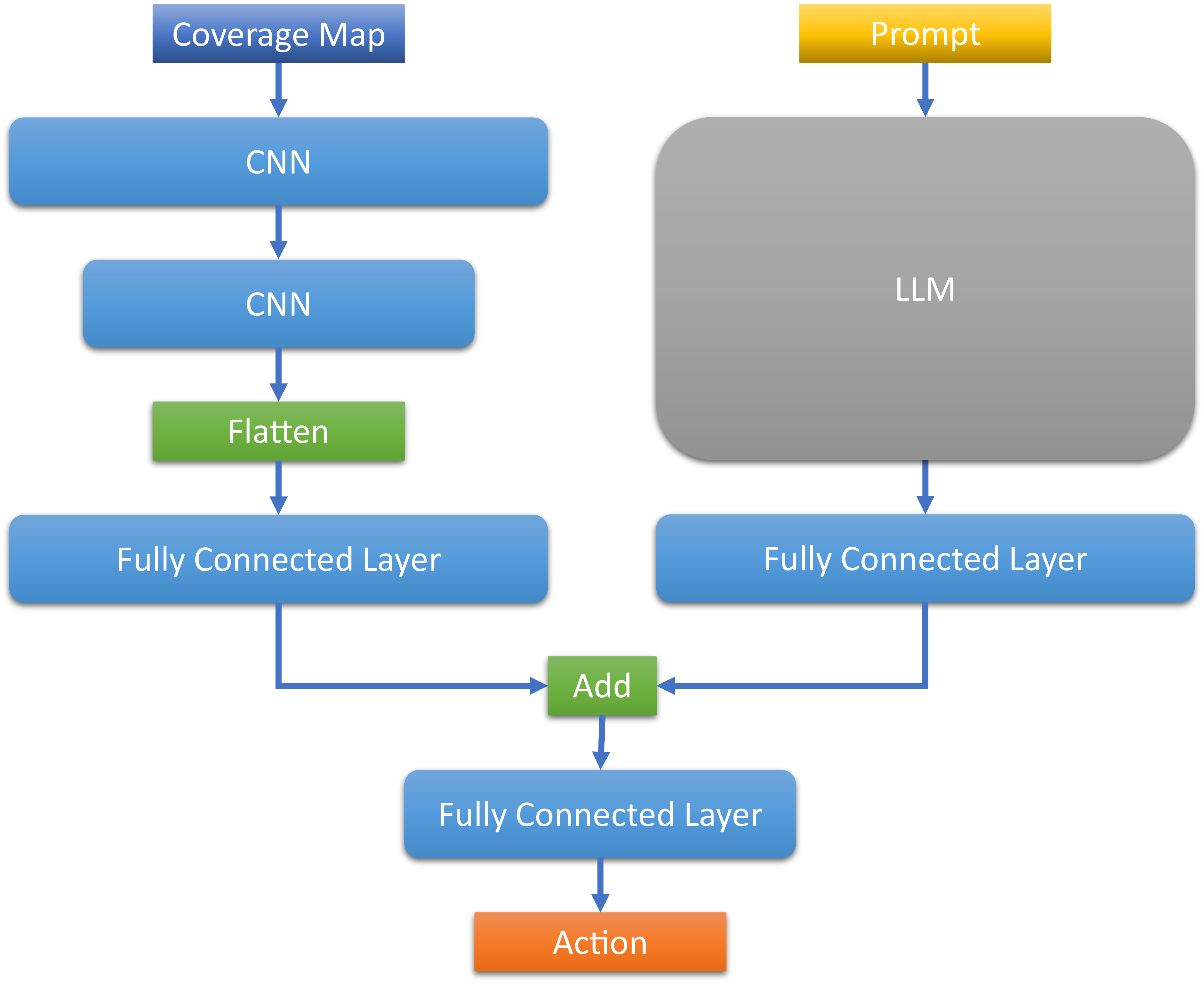}
        \caption{Combined Actor}
    \end{subfigure}
    \caption{Three Actor network architectures.}
    \label{networks}
\end{figure*}

Three variants of the actor network are explored, as shown in Fig. \ref{networks}. The first variant uses a model similar to the critic, employing CNNs to interpret the state from the coverage map, serving as the baseline. The second, an LLM-based model, begins with a text-based prompt processed by the LLM, followed by fully connected layers leading to action selection. This model leverages the pre-trained knowledge of the LLM, offering clear advantages in understanding objectives from prompts.

The CNN-based actor model has the capability to interpret detailed numerical data representations of states from scratch. It lacks pre-existing knowledge and must learn to infer objectives solely from the loss function, which can be a rigorous and involved training process. In contrast, the LLM-based actor model, while limited to the textual input it can process and the maximum sequence length it can accommodate, leverages a pre-trained language model rich with encoded information. This pre-training allows it to understand the objectives clearly and immediately from the input prompts, offering a significant advantage in terms of prior knowledge and ease of objective identification.

The third variant combines CNNs and LLMs, processing the coverage map with CNNs and handling textual prompts with an LLM, integrating both data forms to inform decision-making. This hybrid approach optimizes the strengths of both model types, enhancing decision accuracy and performance in network deployment scenarios.

The choice of the reward function is pivotal for agents to make effective and accurate decisions. In the context of the wireless communication scenario described, the reward function should be closely linked to the received signal power of the users. Consequently, the reward function can be expressed as follows:

\begin{equation}
    r(s,a) = \sum_{i=1}^{16}\alpha_i \times k_i\ \text{dB},
\end{equation}
where $k_i$ represents the received signal power for User \textit{i}, and $\alpha_i$ is the coefficient that modulates the influence of User \textit{i}'s signal power on the reward. This formulation ensures that the reward is directly proportional to the critical performance metric of signal strength, facilitating targeted and efficient training of the agent.

Prompt selection is extremely important to effectively harness LLMs' power. And since we use an RL based strategy to train LLMs, it is even more crucial to have a suitable prompt to describe the state to the LLMs appropriately. To this end, we appealed to the help of another strong language model, none other than ChatGPT. The main purpose of the prompt is to clearly describe the state to the LLMs. Furthermore, to maximize the understanding of the prompt's linguistic features, we aimed for ChatGPT to provide a clear and concise description of the main objective. This approach ensures that, unlike the first variant, LLM-based actor models do not solely rely on the loss function to infer the objective. After describing the format of the prompt and the limitation of maximum sequence length, we asked ChatGPT to come up with a pool of prompts, each of which we later empirically evaluated to detect the best one. The final prompts that we used throughout the experiments are in the following format:

\begin{quote}
    Task: Optimize base station positioning and orientation to cater to users based on their 3D locations. Objective: Maximize signal strengths for all the users. User details provided: User 1 at position $(x_1,y_1,z_1)$ gets $k_1$ dB, ..., User 16 at position $(x_{16},y_{16},z_{16})$ gets $k_{16}$dB. Utilize this data to configure the base station for optimal signal distribution.
\end{quote}

\section{Experiments \& Results}
\label{exp}
We trained three actor configurations and the critic network using the DDPG algorithm across three scenarios, with setups illustrated in Figs. \ref{networks} and \ref{critic}. Key training parameters included a learning rate of $5\times10^{-4}$ for actors and $5\times10^{-3}$ for the critic, a target network update rate of $10^{-3}$, a batch size of 8, and a hidden layer size of 256 for all networks. All the parameters are chosen based on empirical findings. Training was conducted on a GeForce RTX 4090 GPU, completing 1000 steps for each model and monitoring convergence.

A total of 1000 steps are taken for the trainings of all models and the convergence of the models are examined. Every 100 training steps, we paused to perform 20 evaluation steps, during which neural network parameters were frozen and noise factors, outlined in Section \ref{ddpg}, were removed to yield deterministic actions. Post-evaluation, we averaged the rewards from these steps to gauge the policy's effectiveness at that training stage, using 5 random seeds to ensure consistency. Rewards were calculated as the average received signal power in dB for all users, adjusted by a factor of $1/40$.

\begin{equation}
    r(s,a) = \frac{1}{16}\sum_{i=1}^{16}\frac{k_i}{40}\ dB.
\end{equation}

\begin{figure*}
    \centering
    \begin{subfigure}{0.33\textwidth}
        \centering
        \includegraphics[width=\textwidth]{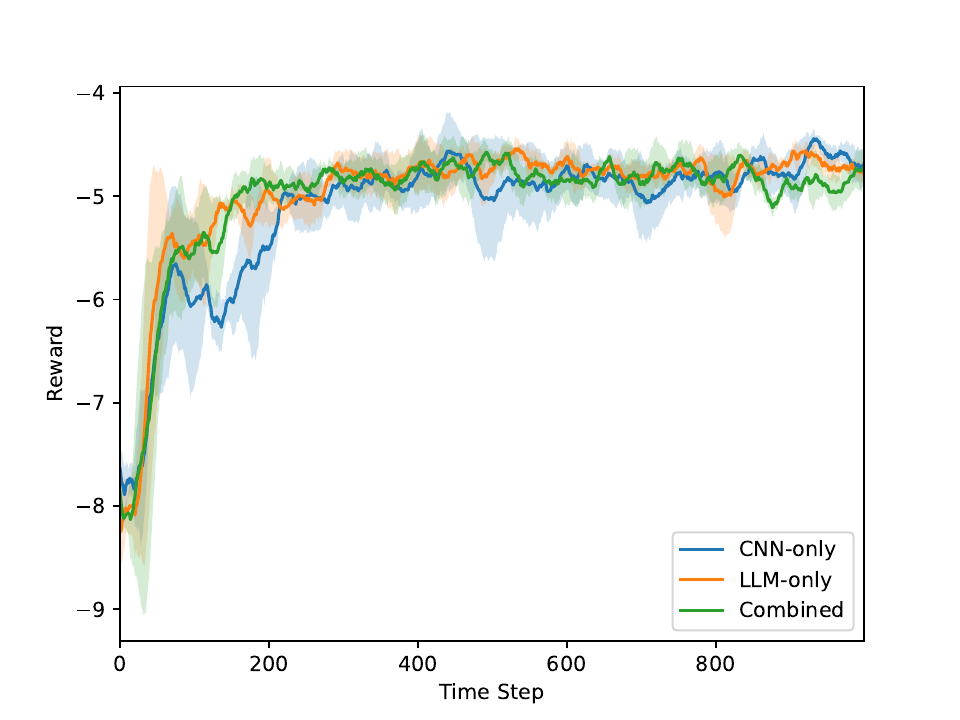}
        \caption{Case 1}
        \label{training-rewards-a}
    \end{subfigure}%
    ~
    \begin{subfigure}{0.33\textwidth}
        \centering
        \includegraphics[width=\textwidth]{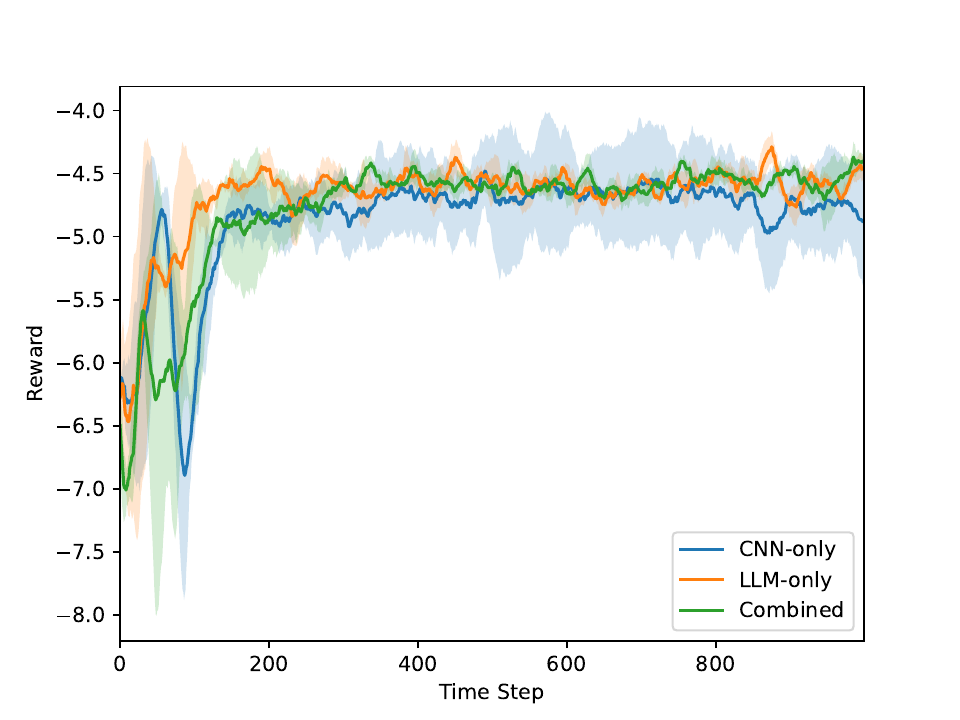}
        \caption{Case 2}
        \label{training-rewards-b}
    \end{subfigure}%
    ~
    \begin{subfigure}{0.33\textwidth}
        \centering
        \includegraphics[width=\textwidth]{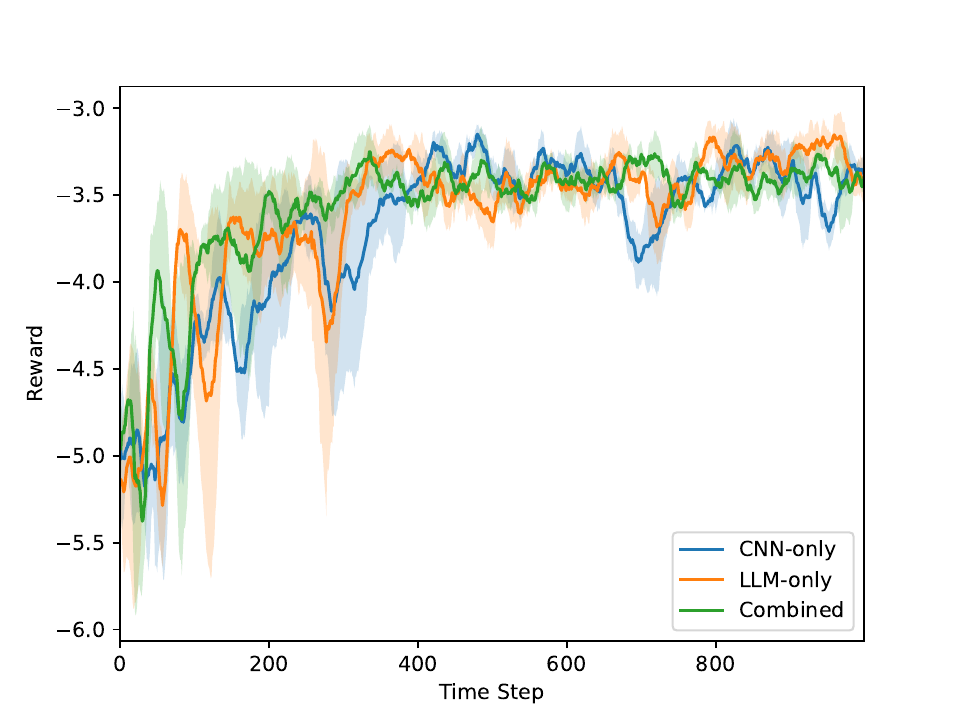}
        \caption{Case 3}
        \label{training-rewards-c}
    \end{subfigure}
    \caption{Comparative training reward dynamics over time for three models. The shaded regions represent the variance in reward values at each time step based over five random seeds.}
    \label{train_rewards}
\end{figure*}

The scaling factor is empirically chosen to stabilize the training.
The rewards gathered during the training across time steps by the agents for all actor settings in three different scenarios are given in Fig. \ref{train_rewards}. The plots are smoothed with a window size of 50. 

Training results indicate that actor networks utilizing LLMs achieve faster and more consistent convergence. The mean performance over five seeds is shown as solid lines, with shades representing standard deviation. As depicted in Fig. \ref{training-rewards-a}, both \textit{LLM-Only} and \textit{Combined} actors converge approximately 150 time steps faster than the \textit{CNN-Only} actor. While all models show consistent performance post-convergence in Case-1, LLM-assisted models converge quicker, with the \textit{Combined} actor slightly outperforming the \textit{LLM-Only} actor until convergence, though the difference in performance is minor.

In Case-2, shown in Fig. \ref{training-rewards-b}, LLM-assisted models (\textit{LLM-Only} and \textit{Combined}) demonstrate notably quicker and more stable convergence across time steps and seeds compared to the \textit{CNN-Only} actor, whose performance shows significant variability. The \textit{LLM-Only} model is initially more stable, maintaining higher performance for about 300 steps, but eventually, both LLM-assisted models level off to similar performance metrics.

In the scenario depicted in Fig. \ref{training-rewards-c} for Case-3, all models exhibit significant fluctuations in their training curves throughout the session. This variability is attributed to the users being more dispersed across the map compared to other cases. Although not immediately apparent, a detailed examination reveals that the curve for the \textit{Combined} actor is comparatively the most stable. This observation suggests that the \textit{Combined} actor may be better at handling the challenges posed by user dispersion in the training environment.

As discussed earlier in this section, we implemented an evaluation process in which we omitted the noise factor that is typically added to the actions determined by the agent during training. We believe that analyzing the agents' performances based on their actual outputs, without the influence of noise, provides a more accurate indicator of their true performance capabilities. This approach allows for a clearer assessment of the effectiveness of the agents' decision-making processes.

\begin{figure*}
    \centering
    \begin{subfigure}{0.33\textwidth}
        \centering
        \includegraphics[width=\textwidth]{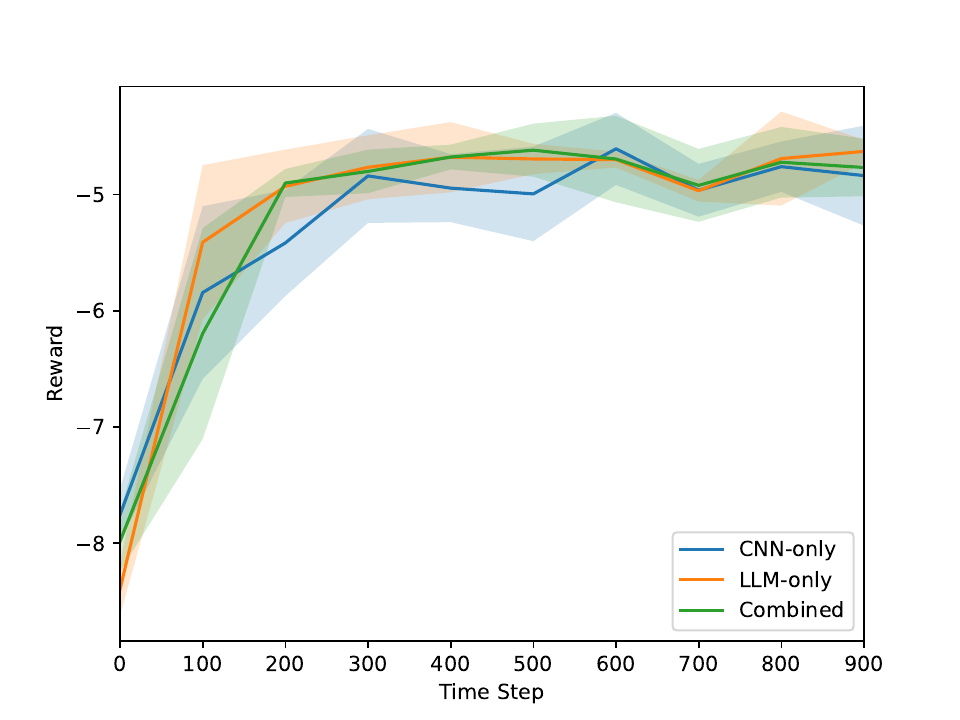}
        \caption{Case 1}
        \label{eval-rewards-a}
    \end{subfigure}%
    ~
    \begin{subfigure}{0.33\textwidth}
        \centering
        \includegraphics[width=\textwidth]{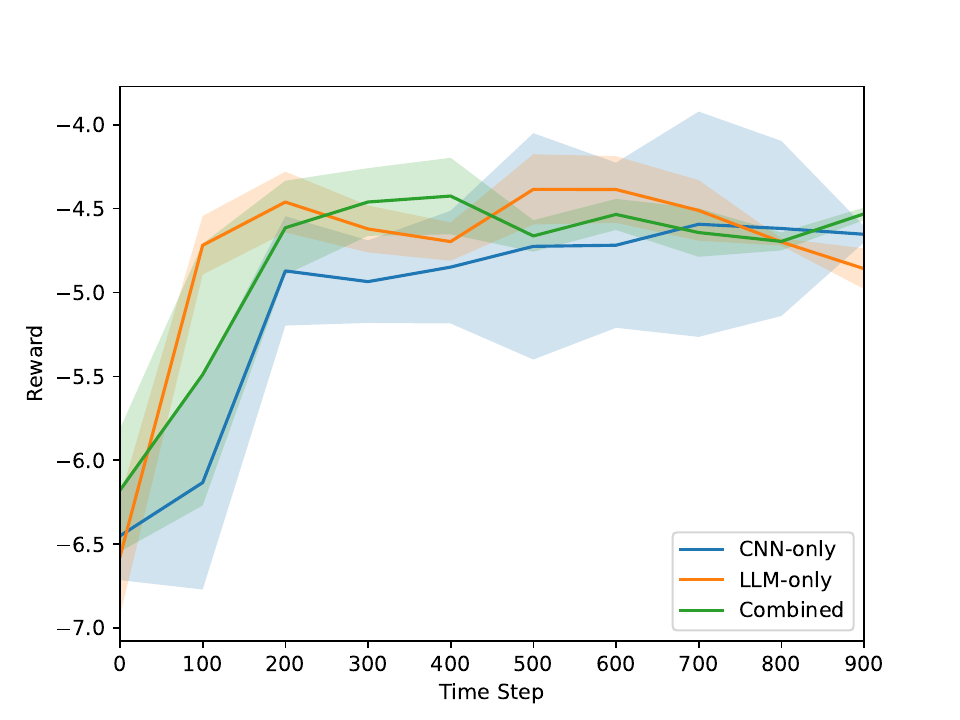}
        \caption{Case 2}
        \label{eval-rewards-b}
    \end{subfigure}%
    ~
    \begin{subfigure}{0.33\textwidth}
        \centering
        \includegraphics[width=\textwidth]{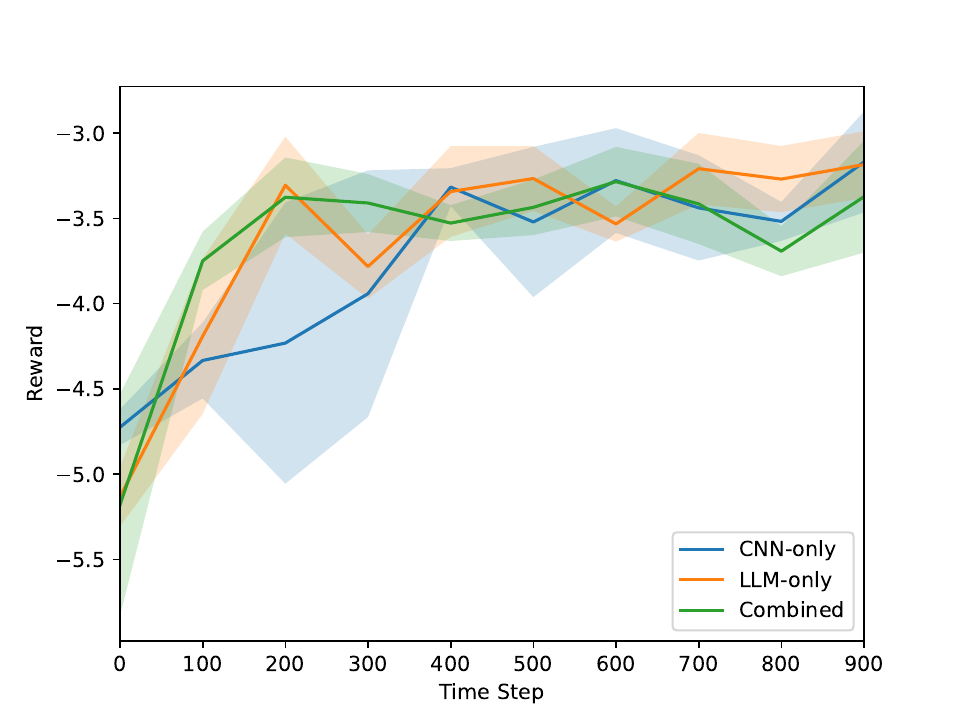}
        \caption{Case 3}
        \label{eval-rewards-c}
    \end{subfigure}
    \caption{Comparative evaluation reward dynamics over time for three models. The shaded regions represent the variance in reward values at each time step based over five random seeds.}
    \label{eval_rewards}
\end{figure*}

The evaluation occurs at specific intervals during the training, specifically every 100 steps. During each evaluation phase, 20 steps are taken, and the total reward accumulated is then averaged over these 20 evaluation steps. The results of these evaluations are presented in Fig. \ref{eval_rewards}, providing insights into the performance of the agents at these intervals. This method helps in understanding the effectiveness of the training at different stages.

Analyzing Fig. \ref{eval-rewards-a}, it appears that all models perform comparably in Case-1, indicating that each actor model could effectively identify the optimal location and orientation to optimally serve users. This uniformity in performance suggests that the scenario presented in Case-1 might not be challenging enough to differentiate the capabilities of the various models.

Conversely, Fig. \ref{eval-rewards-b} reveals that LLM-assisted models surpass the \textit{CNN-Only} actor in terms of both convergence speed and performance stability. The mean curves for the LLM-assisted models consistently remain above that of the \textit{CNN-Only} actor, and exhibit significantly lower variance. While the LLM-assisted actors generally demonstrate similar performance levels, the \textit{LLM-Only} actor achieves convergence one evaluation cycle, or 100 training steps, earlier than the \textit{Combined} actor. Aside from this slight difference in the rate of convergence, both LLM-assisted actors exhibit stable performance throughout the training.

In the context of Case-3, as depicted in Fig. \ref{eval-rewards-c}, there is a notable difference in the convergence rates between the LLM-assisted actors and the \textit{CNN-Only} actor. Although this distinction was not apparent in their training curves shown in Fig. \ref{training-rewards-c}, the evaluation process highlights that the LLM-assisted actors reach convergence three evaluation cycles, or 300 training steps, earlier than the \textit{CNN-Only} actor. This demonstrates the enhanced efficiency of the LLM-assisted models during the evaluation phases, despite similar performances during training. Furthermore, among the LLM-assisted actors, their performance levels are comparable, and they achieve convergence at the same evaluation cycle.

We can also consider the results as an ablation study perspective where in the main model, i.e., \textit{Combined}, we utilize both CNN and LLMs, and we omit CNN for \textit{LLM-Only} actor and omit LLM for \textit{CNN-Only} actor. In all scenarios, we observed that omitting CNN did not affected the performance significantly where omitting LLMs strikingly lowered the performance.

\section{Conclusion}
\label{conc}
In this paper, we suggested a novel methodology to exploit LLMs for wireless network deployment task. We argue that our contribution is a significant milestone in integrating the capabilities of LLMs into the field of wireless communications, given that research in this area is notably limited and existing studies often lack a robust methodology accompanied by realistic experiments.

Our results suggest that the methodology we proposed is successful to utilize the LLMs' power in the devised problem as LLM-assisted models outperformed the model that does not use LLMs where even performing on par would have been a significant outcome considering the sparsity of studies on using LLMs in wireless domain. We believe proposing such a method with the successful results highlights the under-explored potential of using LLMs in the wireless communications domain.

\section*{Acknowledgements}
This material is based upon work supported in part by the U.S. Department of Energy, Office of Science, Office of Advanced Scientific Computing Research, Early Career Research Program under Award Number DE-SC-0023957, and in part by the National Science Foundation under Grant No. 2323300.

\bibliographystyle{IEEEtran}
\bibliography{references}

\begin{thebibliography}{10}
\providecommand{\url}[1]{#1}
\csname url@samestyle\endcsname
\providecommand{\newblock}{\relax}
\providecommand{\bibinfo}[2]{#2}
\providecommand{\BIBentrySTDinterwordspacing}{\spaceskip=0pt\relax}
\providecommand{\BIBentryALTinterwordstretchfactor}{4}
\providecommand{\BIBentryALTinterwordspacing}{\spaceskip=\fontdimen2\font plus
\BIBentryALTinterwordstretchfactor\fontdimen3\font minus \fontdimen4\font\relax}
\providecommand{\BIBforeignlanguage}[2]{{%
\expandafter\ifx\csname l@#1\endcsname\relax
\typeout{** WARNING: IEEEtran.bst: No hyphenation pattern has been}%
\typeout{** loaded for the language `#1'. Using the pattern for}%
\typeout{** the default language instead.}%
\else
\language=\csname l@#1\endcsname
\fi
#2}}
\providecommand{\BIBdecl}{\relax}
\BIBdecl

\bibitem{brown2020language}
T.~B. Brown, B.~Mann, N.~Ryder, M.~Subbiah, J.~Kaplan, P.~Dhariwal, A.~Neelakantan, P.~Shyam, G.~Sastry, A.~Askell, S.~Agarwal, A.~Herbert-Voss, G.~Krueger, T.~Henighan, R.~Child, A.~Ramesh, D.~M. Ziegler, J.~Wu, C.~Winter, C.~Hesse, M.~Chen, E.~Sigler, M.~Litwin, S.~Gray, B.~Chess, J.~Clark, C.~Berner, S.~McCandlish, A.~Radford, I.~Sutskever, and D.~Amodei, ``Language models are few-shot learners,'' 2020.

\bibitem{Devlin2018}
J.~Devlin, M.-W. Chang, K.~Lee, and K.~Toutanova, ``Bert: Pre-training of deep bidirectional transformers for language understanding,'' 2018.

\bibitem{sanh2019distilbert}
V.~Sanh, L.~Debut, J.~Chaumond, and T.~Wolf, ``Distil{BERT}, a distilled version of {BERT}: smaller, faster, cheaper and lighter,'' \emph{arXiv preprint arXiv:1910.01108}, 2019.

\bibitem{Goldsmith2005}
A.~Goldsmith, \emph{Wireless Communications}.\hskip 1em plus 0.5em minus 0.4em\relax Cambridge University Press, 2005.

\bibitem{Rappaport2002}
T.~S. Rappaport, \emph{Wireless Communications: Principles and Practice}, 2nd~ed.\hskip 1em plus 0.5em minus 0.4em\relax Prentice Hall, 2002.

\bibitem{sonsurvey}
O.~G. Aliu, A.~Imran, M.~A. Imran, and B.~Evans, ``A survey of self organisation in future cellular networks,'' \emph{IEEE Communications Surveys \& Tutorials}, vol.~15, no.~1, pp. 336--361, 2013.

\bibitem{mullany05son}
F.~J. Mullany, L.~T.~W. Ho, L.~G. Samuel, and H.~Claussen, ``Self-deployment, self-configuration:critical future paradigms for wireless access networks,'' in \emph{Autonomic Communication}, M.~Smirnov, Ed.\hskip 1em plus 0.5em minus 0.4em\relax Berlin, Heidelberg: Springer Berlin Heidelberg, 2005, pp. 58--68.

\bibitem{aliu13son}
O.~G. Aliu, A.~Imran, M.~A. Imran, and B.~Evans, ``A survey of self organisation in future cellular networks,'' \emph{IEEE Communications Surveys \& Tutorials}, vol.~15, no.~1, pp. 336--361, 2013.

\bibitem{ho03son}
L.~T.~W. Ho, L.~G. Samuel, and J.~M. Pitts, ``Applying emergent self-organizing behavior for the coordination of 4g networks using complexity metrics,'' \emph{Bell Labs Technical Journal}, vol.~8, no.~1, pp. 5--25, 2003.

\bibitem{lillicrap2015continuous}
T.~P. Lillicrap, J.~J. Hunt, A.~Pritzel, N.~Heess, T.~Erez, Y.~Tassa, D.~Silver, and D.~Wierstra, ``Continuous control with deep reinforcement learning,'' \emph{arXiv preprint arXiv:1509.02971}, 2015.

\bibitem{roberta}
\BIBentryALTinterwordspacing
Y.~Liu, M.~Ott, N.~Goyal, J.~Du, M.~Joshi, D.~Chen, O.~Levy, M.~Lewis, L.~Zettlemoyer, and V.~Stoyanov, ``Roberta: {A} robustly optimized {BERT} pretraining approach,'' \emph{CoRR}, vol. abs/1907.11692, 2019. [Online]. Available: \url{http://arxiv.org/abs/1907.11692}
\BIBentrySTDinterwordspacing

\bibitem{sionna}
J.~Hoydis, S.~Cammerer, F.~{Ait Aoudia}, A.~Vem, N.~Binder, G.~Marcus, and A.~Keller, ``Sionna: An open-source library for next-generation physical layer research,'' \emph{arXiv preprint}, Mar. 2022.

\bibitem{ornstein}
\BIBentryALTinterwordspacing
G.~E. Uhlenbeck and L.~S. Ornstein, ``On the theory of the brownian motion,'' \emph{Phys. Rev.}, vol.~36, pp. 823--841, Sep 1930. [Online]. Available: \url{https://link.aps.org/doi/10.1103/PhysRev.36.823}
\BIBentrySTDinterwordspacing

\end{thebibliography}

\end{document}